# Recognition of Handwritten Roman Script Using Tesseract Open source OCR Engine

Sandip Rakshit [1], Subhadip Basu [2,#]

[1] Techno India College of Technology, Kolkata, India
[2] Computer Science and Engineering Department, Jadavpur University, India

[#] Corresponding author. E-mail: subhadip@ieee.org

## Abstract

*In the present work, we have used Tesseract 2.01 open source Optical Character Recognition (OCR) Engine under Apache License 2.0 for recognition of handwriting samples of lower case Roman script. Handwritten isolated and free-flow text samples were collected from multiple users. Tesseract is trained to recognize user-specific handwriting samples of both the categories of document pages. On a single user model, the system is trained with 1844 isolated handwritten characters and the performance is tested on 1133 characters, taken form the test set. The overall character-level accuracy of the system is observed as 83.5%. The system fails to segment 5.56% characters and erroneously classifies 10.94% characters.*

## 1. Introduction

*Optical Character Recognition (OCR)* systems ease the barrier of the keyboard interface between man & machine to a great extent, and help in office automation with huge saving of time and human effort. Such systems allow desired manipulation of the scanned text as the output is coded with ASCII or some other character code from the paper based input text. For a specific language based on some alphabet, OCR techniques are either aimed at printed text or handwritten text. The present work is aimed at the later.

Machine recognition of handwritten text is one of the challenging areas of research for the pattern recognition community. In general, OCR systems have potential applications in extracting data from filled in forms, interpreting handwritten addresses from postal documents for automatic routing, automatic reading of bank cheques etc. The core component of such application softwares is an OCR engine, equipped with the key functional modules like line extraction, line-to-word segmentation, word-to-character segmentation, character recognition and word-level lexicon analysis using standard dictionaries.

Development of a handwritten OCR engine with high recognition accuracy is a still an open problem for the research community. Lot of research efforts have already been reported [1-8] on different key aspects of handwritten character recognition systems. In the current work, instead of developing a new handwritten OCR engine from scratch, we have used Tesseract 2.01 [9], an open source OCR Engine under Apache License 2.0, for recognition of handwritten pages consisting of lower case characters of Roman script. Tesseract OCR engine provides high level of character recognition accuracy on poorly printed or poorly copied dense text. But the performance of this OCR engine is not extensively tested on recognition of handwritten characters. This has been one of the major motivations behind the current work, presented in this paper.

In the current work, we have used Tesseract to perform user specific training on handwriting samples of both isolated and free-flow texts, written using lower case Roman script. The performance is evaluated on both the categories of document pages for observation of character level and word level accuracies.

## 2. Overview of the Tesseract OCR engine

Tesseract is an open source (under Apache License 2.0) offline optical character recognition engine, originally developed at Hewlett Packard from 1984 to 1994. Tesseract was first started as a PhD research project in HPLabs, Bristol [10]. In the year 1995 it is sent to UNLV where it proved its worth against the commercial engines of the time [11]. In the year 2005 Hewlett Packard and University of Nevada, Las Vegas, released it. Now it is partially funded by Google [12] and released under the Apache license, version 2.0.

The latest version, Tesseract 2.03 is released in April, 2008.

Like any standard OCR engine, Tesseract is developed on top of the key functional modules like, line and word finder, word recognizer, static character classifier, linguistic analyzer and an adaptive classifier. However, it does not support document layout analysis, output formatting and graphical user interface. Currently, Tesseract can recognize printed text written in English, Spanish, French, Italian, Dutch, German and various other languages.

To train Tesseract in English language 8 data files are required in tessdata sub directory. The 8 files used for English are to be generated as follows:

tessdata/eng.freq-dawg
tessdata/eng.word-dawg
tessdata/eng.user-words
tessdata/eng.inttemp
tessdata/eng.normproto
tessdata/eng.pffmtable
tessdata/eng.unicharset
tessdata/eng.DangAmbigs

## 3. The present work

In the present work, we have used Tesseract version 2.01for recognition of handwriting samples of both isolated and free-flow texts, written using lower case Roman script. Key functional modules of the developed system are discussed the following sub-sections.

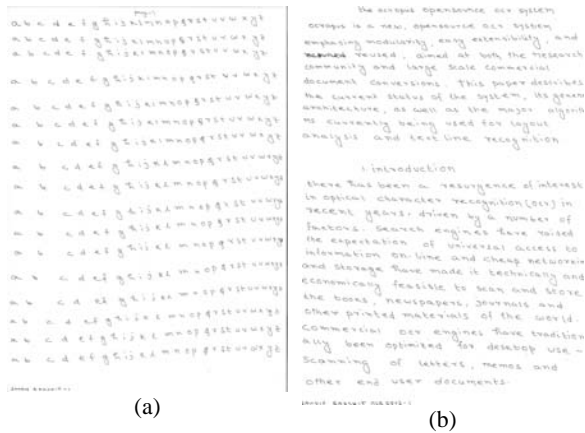

(a)     (b)

**Fig. 1(a-b). Sample document pages containing training sets of isolated characters and free flow text**

**3.1. Collection of the dataset**

For collection of the dataset for the current experiment, we have concentrated on lower case characters of Roman script. Six handwritten document pages were collected from each of the four different users in two types of datasets. In the first set, four pages of isolated handwritten lower case Roman characters were collected, as shown in Fig. 1(a), and in the second set, two pages of free-flow handwritten text, as shown in Fig. 1(b), written from technical articles, were collected from each users. For each user, three pages from the first set and one page from the second dataset were considered for training the Tesseract OCR engine. The remaining two pages, one from each set, constitute the test set for the current experiment.

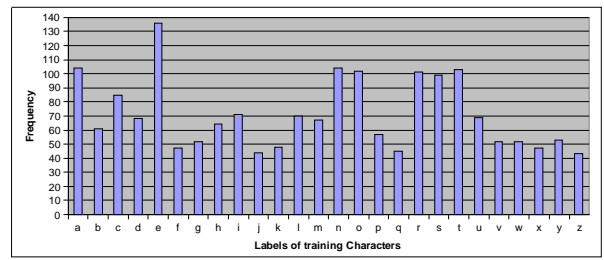

**Fig. 2. Frequency distribution of different character samples during training**

The training dataset contains around 70 sample sets of isolated lower case Roman characters for each user and around 120 words (around 650 characters) of free-flow text. For example, the training set for the first user consists of a varying distribution of 1844 labeled lower case character samples, as shown in Fig. 2.

**3.2. Labeling training data**

For labeling the training samples using Tesseract we have taken help of a tool named bbTesseract [12]. To generate the training files for a specific user, we need to prepare the box files for each training images using the following command:

*tesseract fontfile.tif fontfile batch.nochop makebox*

The box file is a text file that includes the characters in the training image, in order, one per line, with the coordinates of the bounding box around the image. The new Tesseract 2.01 has a mode in which it will output a text file of the required format. Some times the character set is different to its current training, it will naturally have the text incorrect. In that case we have to manually edit the file (using bbTesseract) to correct the incorrect characters in it. Then we have to rename fontfile.txt to fontfile.box. Fig. 3 shows a screenshot of the bbTesseract tool, used for labeling the training set.

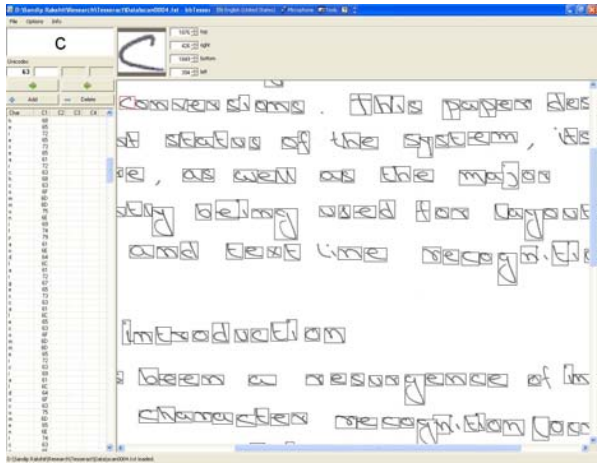

**Fig.3. A sample screenshot of the bbTesseract tool**

**3.3. Training the data using Tesseract OCR engine**

For training a new handwritten character set for any user, we have to put in the effort to get one good box file for a handwritten document page, run the rest of the training process, discussed below, to create a new language set. Then use Tesseract again using the newly created language set to label the rest of the box files corresponding to the remaining training images using the process discussed in section 3.2.

For each of our training image, boxfile pairs, run Tesseract in training mode using the following command:

*tesseract fontfile.tif junk nobatch box.train*

The output of this step is fontfile.tr which contains the features of each character of the training page. The character shape features can be clustered using the mftraining and cntraining programs:

*mftraining fontfile_1.tr fontfile_2.tr ...*

This will output three data files: inttemp , pffmtable and Microfeat, and the following command:

*cntraining fontfile_1.tr fontfile_2.tr ...*

This will output the normproto data file. Now, to generate the unicharset data file, unicharset_extractor program is used as follows:

*unicharset_extractor fontfile_1.box fontfile_2.box ...*

Tesseract uses 3 dictionary files for each language. Two of the files are coded as a Directed Acyclic Word Graph (DAWG), and the other is a plain UTF-8 text file. To make the DAWG dictionary files a wordlist is required for our language. The wordlist is formatted as a UTF-8 text file with one word per line. The corresponding command is:

*wordlist2dawg frequent_words_list freq-dawg*
*wordlist2dawg words_list word-dawg*

The third dictionary file name is user-words and is usually empty. The final data file of Tesseract is DangAmbigs file. This file cannot be used to translate characters from one set to another. The DangAmbigs file may be empty also.

Now we have to collect all the 8 files and rename them with a lang. prefix, where lang is the 3-letter code for our language and put them in our tessdata directory. Tesseract can then recognize text in our language using the command:

*tesseract image.tif output -l lang*

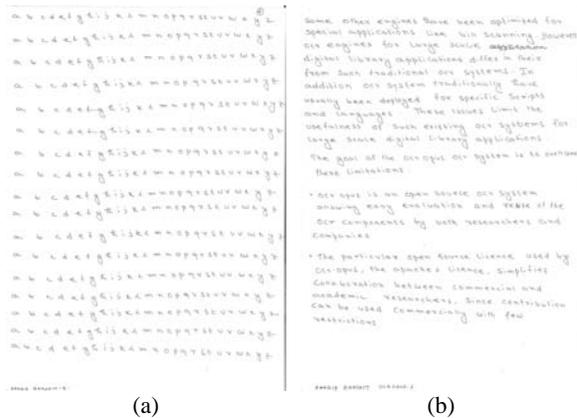

(a)  (b)

**Fig.4. Snapshots of the test pages used for the current experiment**

## 4. Experimental results

For conducting the current experiment, we have considered a single user model with 1844 training samples and 1133 test samples of isolated lower case characters of Roman script. The test pages used for this experiment are shown in Fig. 4. The experiment was focused on testing the core recognition accuracy of Tesseract OCR engine on handwritten document pages. For this purpose, the linguistic analysis module of Tesseract, involving the language files freq-dawg, word-dawg, user-words and DangAmbigs are purposefully left blank.

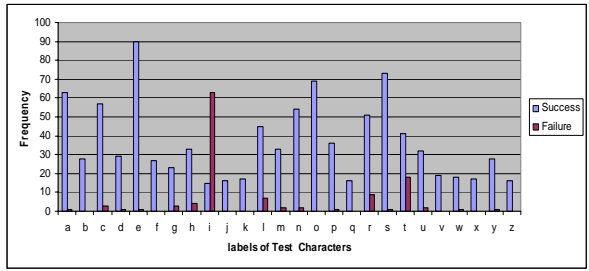

**Fig. 5. Distribution of success and failure cases over the free flow test page.**

The performance of the developed system is evaluated on two datasets, as discussed in section 3.1. Table 1 shows an analysis of both segmentation and recognition performances of the present technique on the test pages. Fig. 5 shows a character wise distribution of success and failure accuracies on the overall test dataset. As observed from the experimentation a significant proportion of the error cases evolve out of the word segmentation failures. This is so because Tesseract is originally designed to recognize printed document pages with uniformity in baseline and character/word spacings. Another source of error is due to the internal segmentation of some of the characters. More specifically, the character 'i' often gets internally segmented into two parts, leading to high individual error rates. This single character alone contributes to around 53% of the overall misclassified cases.

**Table 1. Analysis of recognition performance of the developed system**

|  | Isolated Characters | Free flow Text | | Total Characters |
|---|---|---|---|---|
|  |  | Characters | Words |  |
| Training Set | 1185 | 659 | 137 | 1844 |
| Test Set | 442 | 691 | 120 | 1133 |
| Successful Recognition in Test Set | 89.59% | 79.60% | 45.83% | 83.50% |
| Segmentation Failure | 6.56% | 15.48% | 5.84% | 5.56% |
| Misclassification | 3.85% | 4.92% | 48.33% | 10.94% |

As shown in Table 1, the overall character-level recognition accuracy of the developed system is around 83.5%. The reason behind low word-level accuracy of 45.83% is over-segmentation of some of the constituent characters. The word-level accuracy can further be improved with inclusion of lexicon analyser module of Tesseract. Some of the sample word images successfully segmented and recognized by Tesseract are shown in Fig. 6(a-d). Fig. 7(a-b) shows some of the word images with erroneous segmentation and recognition results.

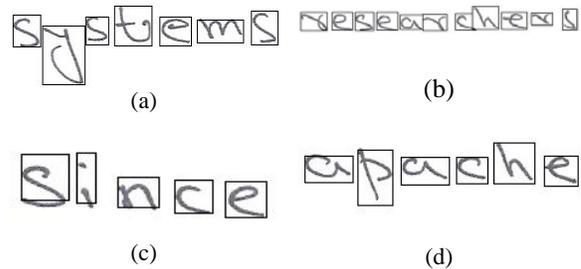

(a)          (b)

(c)          (d)

**Fig. 6(a-d). Some of the successfully segmented and recognized word images.**

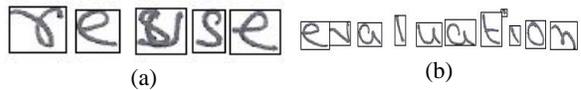

(a)          (b)

**Fig. 7. Some of the misclassified word images**
**(a) Recognition error in the 3$^{rd}$ character**
**(b) Internal segmentation in the 8$^{th}$ character**

## 5. Conclusion

As observed from the experimental results, Tesseract OCR engine fares reasonably with respect to the core recognition accuracy on user-specific handwritten samples of isolated / free-flow text, written using lower-case Roman script. The performance of the system need to be validated on a multi-user platform. A major drawback of the current technique is its failure to avoid over-segmentation in some of the characters. Also the system fails to segment cursive words in many cases. The performance of the designed system may be improved by incorporating more training samples per user and inclusion of word-level dictionary matching techniques.

## 6. Acknowledgements

Technology for necessary supports during the research work. Dr. Subhadip Basu is thankful to the "Center for Microprocessor Application for Training Education and Research", "Project on Storage Retrieval and Understanding of Video for Multimedia" of Computer Science & Engineering Department, Jadavpur University, for providing infrastructure facilities during progress of the work.